\documentclass[runningheads]{llncs}

\usepackage{eccv}

\usepackage{eccvabbrv}
\usepackage{enumitem}
\usepackage{graphicx}
\usepackage{booktabs}
\usepackage{pifont}
\usepackage[accsupp]{axessibility}  

\usepackage{hyperref}
\hypersetup{
colorlinks=true,
linkcolor=black
}

\usepackage{orcidlink}
\usepackage{graphicx}
\usepackage{booktabs}
\usepackage{multirow}
\usepackage{rotating}
\usepackage{makecell}

\begin{document}

\title{Adaptive High-Frequency Transformer for Diverse Wildlife Re-Identification} 

\titlerunning{Adaptive High-Frequency Transformer}


\author{Chenyue Li$^*$\orcidlink{0009-0003-0352-2576} , Shuoyi Chen$^*$\orcidlink{0000-0001-7507-3309},
Mang Ye$^\dagger$\orcidlink{0000-0003-3989-7655}
}

\institute{
National Engineering Research Center for Multimedia Software,\\ Institute of Artificial Intelligence, School of Computer Science, \\Hubei Luojia Laboratory, Wuhan University, Wuhan, China\\
\email{\{chenyueli, chenshuoyi, yemang\}@whu.edu.cn}}
\authorrunning{C.~Li et al.}



\maketitle

\renewcommand{\thefootnote}{}
\footnotetext{$^*$ Equal contributions. $^\dagger$ Corresponding author.}
\renewcommand{\thefootnote}{\arabic{footnote}}

\begin{abstract}
  Wildlife ReID involves utilizing visual technology to identify specific individuals of wild animals in different scenarios, holding significant importance for wildlife conservation, ecological research, and environmental monitoring. Existing wildlife ReID methods are predominantly tailored to specific species, exhibiting limited applicability. Although some approaches leverage extensively studied person ReID techniques, they struggle to address the unique challenges posed by wildlife. Therefore, in this paper, we present a unified, multi-species general framework for wildlife ReID. Given that high-frequency information is a consistent representation of unique features in various species, significantly aiding in identifying contours and details such as fur textures, we propose the Adaptive High-Frequency Transformer model with the goal of enhancing high-frequency information learning. To mitigate the inevitable high-frequency interference in the wilderness environment, we introduce an object-aware high-frequency selection strategy to adaptively capture more valuable high-frequency components. Notably, we unify the experimental settings of multiple wildlife datasets for ReID, achieving superior performance over state-of-the-art ReID methods. In domain generalization scenarios, our approach demonstrates robust generalization to unknown species. Code is available at \url{https://github.com/JigglypuffStitch/AdaFreq.git}.
  
  \keywords{Wildlife Re-Identification \and Transformer \and High-Frequency}
\end{abstract}

\section{Introduction}\par
\label{sec:intro}

Wildlife Re-Identification (ReID) aims to accurately identify specific individual animals in images or videos captured at different time points or locations \cite{ye2024transformer, papafitsoros2022seaturtleid, li2019atrw, zhang2021yakreid}. In contrast to regular animal classification, wildlife ReID necessitates a more advanced level of differentiation among individuals within the same species. This task is crucial for monitoring the living conditions, migratory habits, and reproductive situations of targeted wild animals, with broad applications in the conservation of endangered species, ecological research, and livestock management. Unlike person ReID \cite{ye2020augmentation, ye2023channel, chen2023sketchtrans} task, wildlife ReID necessitates the processing of various species, each with its own unique appearance and behavioral patterns. This requires technologies to possess substantial generalizability and adaptability.

In the realm of wildlife ReID, the majority of efforts are directed towards a specific type of animal \cite{nepovinnykh2022sealid, li2019atrw, wang2021giant, korschens2019elpephants, zhang2021yakreid}. This indicates that in practical applications, the need to design separate methods for each species severely hampers universality and efficiency. In recent years, researchers have explored the re-identification of a diverse range of wildlife in domain generalization scenarios \cite{jiao2024toward}. They construct a large-scale, multi-species dataset to acquire additional knowledge, thereby enhancing the generalization capability for identifying unknown species. However, for tasks demanding highly refined recognition, such as wildlife ReID, this approach proves challenging to achieve the desired level of accuracy. Furthermore, the acquisition of large-scale wildlife data is not easily attainable, and the inability to encompass a sufficiently diverse range of animal data further limits generalization capabilities. Therefore, it is crucial to formulate a unified ReID method applicable to multiple species and capable of meeting refined and superior performance in specific scenarios.

\begin{figure}[t]
    \centering
    \includegraphics[width=\linewidth]{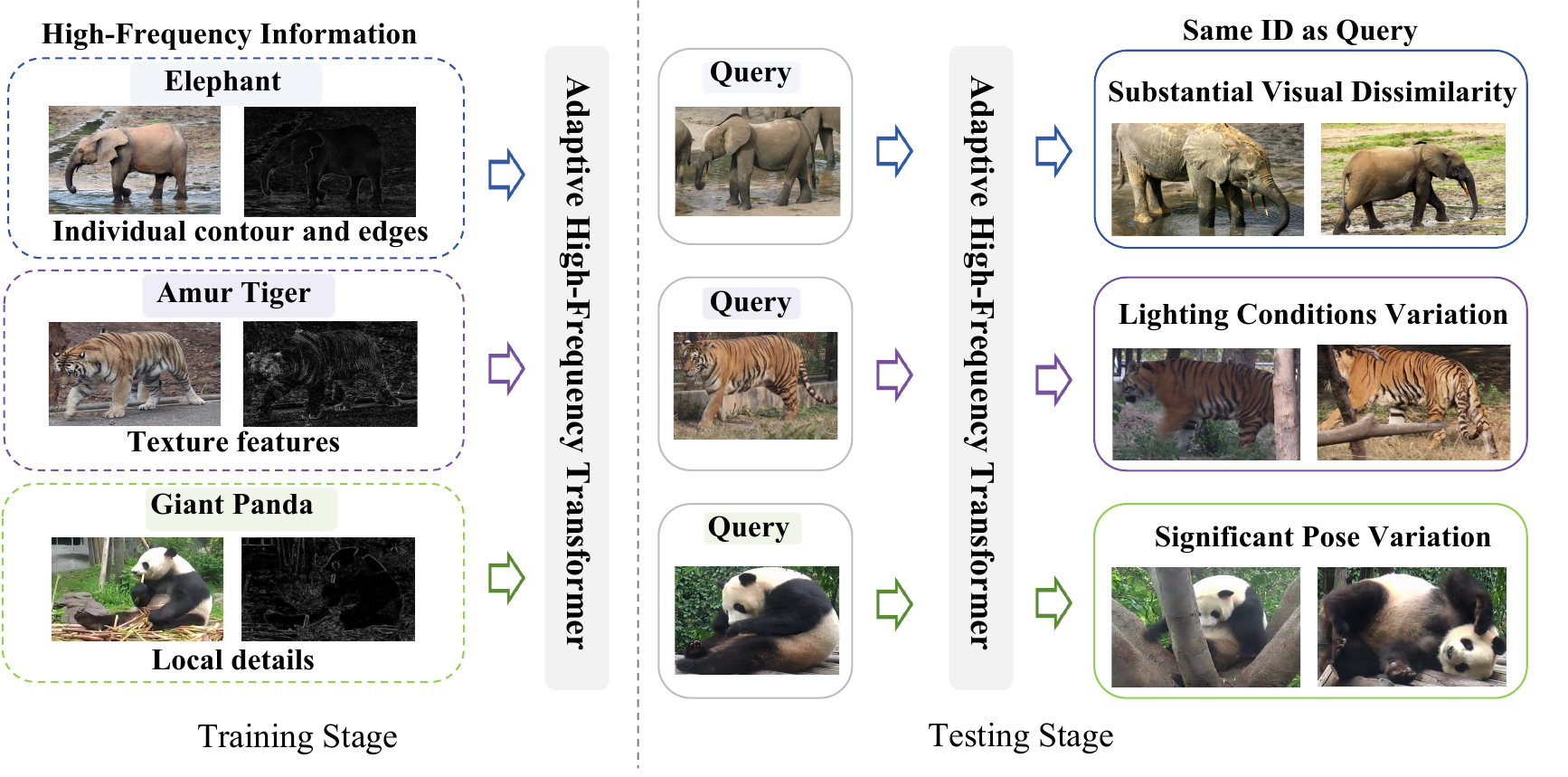}
    \caption{By capturing discriminative features such as texture, contour, and fine details, high-frequency information displays unique features specific to each wild species, playing a crucial role in universal wildlife re-identification. 
    } 
    \label{intro}
\end{figure}

Existing methods are typically designed for specific species, focusing on the extraction of particular physical traits unique to that species for subsequent matching. This involves features such as the edges of humpback whale flukes \cite{weideman2020extracting} or the pelage pattern of ringed seals \cite{nepovinnykh2022matching}.
However, these methods rely on the inherent characteristics of the species, posing challenges to the applicability to other wildlife. Furthermore, several studies employ well-established convolutional neural network(CNN) based methods to acquire discriminative representations \cite{wang2021giant,bouma2018individual,bruslund2020re}. These methods draw inspiration from person ReID but fail to consider the unique challenges of wildlife. For humans, distinguishable features include facial details, hairstyles, clothing, and accessories. When considering key features of wildlife, they are typically manifested in the texture of fur or scales, patterns or spots, and the contour of edges. Such characteristics are uniformly reflected in the high-frequency information of images, where high-frequency information refers to features exhibiting rapid changes or fine structures in the image \cite{lin2023deep}, as illustrated in Fig.\ref{intro}. Therefore, we believe that high-frequency information plays a key role in unifying the ReID of different wild animals. A direct method is to extract the high-frequency information from the image for augmentation. However, in natural environments, wildlife images often have more complex and diverse backgrounds and most data lack clear bounding boxes. Considerable environmental noise interference is also included in high-frequency information, such as textures from leaves and grass.

To address these challenges, in this paper we propose an Adaptive High-Frequency Transformer, offering a universal framework for wildlife ReID applicable to various species. With the core objective of improving discriminative feature learning, we design three strategies to precisely direct high-frequency information. Considering the instability of high-frequency details such as fur patterns and small spots in wildlife due to variations in lighting and posture, we propose a new frequency-domain mixed data augmentation method to enhance the robustness of the model. Specifically, it blends the high-frequency representation of the image with the frequency-domain representation of the original image. The advantage lies in the frequency-domain level operation, which avoids introducing redundant information. Furthermore, to mitigate high-frequency noise in complex natural environments, we design an object-aware dynamic selection strategy to flexibly capture high-frequency information more relevant to the target. The key idea lies in leveraging the Transformer, where the global attention of the original image can be regarded as a guiding mechanism to selectively filter out tokens with negative interference. Finally, in order to mitigate the risk of excessive emphasis on high-frequency details at the cost of sacrificing original visual information, we design a feature equilibrium loss to constrain the disparity between high-frequency features and global features.

In summary, our proposed approach strives for both the universality of the model across various species and its adaptability among different targets. \textbf{1) From a universal perspective}, high-frequency information acts as a universal key, linking unique features across diverse wildlife species. By intensifying attention on high-frequency information, our model demonstrates the capacity to comprehensively and universally capture distinguishing features of wildlife, transcending limitations associated with specific species. This broadens the applicability of our ReID model.   \textbf{ 2) From an adaptive perspective}, our model takes into account individual variations among different species. By introducing an object-aware adaptive mechanism for high-frequency information, we can better capture the diverse high-frequency features presented by different targets, such as fur textures and pattern shapes. This adaptability contributes to improving the model's accuracy in recognizing individual wildlife, enabling it to flexibly handle various species. 
Notably, we conduct extensive experiments and unify the experimental settings of multiple wildlife datasets for ReID. Results demonstrate that our method significantly outperforms state-of-the-art ReID methods on diverse wildlife datasets, covering terrestrial, aquatic, and aerial species. Additionally, our model trained on large-scale multi-species datasets and evaluated in the domain generalization setting, maintains reliable generalization to unfamiliar species.

\section{Related Works}
\textbf{Wildlife Re-Identification.}Recently, the rapid development of deep learning \cite{10205526, yang2024emollmmultimodalemotionalunderstanding, intclip, FLSurveyandBenchmarkforGenRobFair_TPAMI24, FCCLPlus_TPAMI23, FPL_CVPR23, FCCL_CVPR22} has made convolutional neural networks (CNNs) widely adopted for wildlife re-identification, leveraging them for feature extraction\cite{ahmed2015improved,zhao2014learning, li2014deepreid} and metric learning\cite{liao2015person,bkak2017deep,xiong2014person}. While person ReID technology has reached a mature state, wildlife ReID remains in an early stage. Many of them are specific to particular species, restricting their applicability\cite{matthe2017comparison, norouzzadeh2018automatically, halloran2015applying}.
We can divide them into the following categories.
\textit{(1) Global Feature Learning}: Building on traditional person ReID methods, many approaches use whole animal images to extract distinctive features. Wang et al.\cite{wang2021giant} developed a multi-stream feature fusion network aimed at extracting and integrating both local and global features of giant pandas. For manta ray ReID, where distinctive patterns vary unpredictably, Moskvyak et al. \cite{moskvyak2021robust} proposed a specialized loss to reduce feature distance between different views of the same individual, enhancing pose-invariant feature learning.
Such methods do not require prior knowledge of specific species when designing feature extraction mechanisms, making them particularly suitable for practical applications across a diverse range of species. However, existing methods primarily rely on traditional person ReID technologies.
\textit{(2) Species-specific Feature Extraction}: Some wildlife have distinctive patterns, and these methods crop these specific pattern areas for local identification. These methods are adept at extracting discriminative features from specific body parts in a range of animal species, such as dolphin fins\cite{bouma2018individual,weideman2017integral,konovalov2018individual}, the heads of cattle\cite{bergamini2018multi}, tail features of whales\cite{cheeseman2022advanced}, elephant ears\cite{weideman2020extracting}, the pelage patterns of ringed seals\cite{nepovinnykh2020siamese}, etc. However, in practical applications, significant variations in perspective and partial occlusions often occur, making it challenging to capture clear local features for each individual, resulting in unreliable images. These methods typically face difficulties in being generalized to animals of other species.
\textit{(3) Auxiliary Information Integration}: Li et al.\cite{li2019atrw} leveraged pose key point estimation outcomes to segment tiger images into seven components, facilitating local feature learning. Following this, Zhang et al.\cite{zhang2021yakreid} adopted a simplified pose definition for either side of a yak's head, serving as an auxiliary supervisory signal to improve feature learning. Yet, these approaches require supplementary annotations and depend on disparate auxiliary information specific to various animals, making it challenging to apply a standardized method across multiple species.

\textbf{Transformer Based Object ReID.} 
The extensively studied realms of person and vehicle ReID have long been dominated by CNN-based methods, but the introduction of Vision Transformer has revolutionized this field \cite{ye2024transformer}. For general object ReID, TransReID \cite{he2021transreid}, as a purely ViT-based ReID model, significantly enhances cross-camera object ReID accuracy through local feature learning and viewpoint information. Besides, several studies \cite{li2021diverse, zhang2021hat, li2022pyramidal} focus on hybrid models that merge ViT with CNNs.
A dual cross-attention learning method \cite{zhu2022dual} is proposed to enhance the learning of global and local features by improving the attention mechanism. Some methods \cite{wang2022pose} also utilize auxiliary information like pose estimation to learn more effective human body-related features. While these methods exhibit distinct advantages, they have limitations when dealing with the unique challenges posed by wildlife. For multi-species wildlife ReID, recent endeavors have introduced universal recognition methods for domain generalization \cite{jiao2024toward}. However, they primarily rely on leveraging knowledge from CLIP \cite{radford2021learning} to enhance descriptive information for improved generalization, resulting in limited performance for fine-grained recognition of specific species. In comparison, this paper is more targeted at wildlife features under a unified model, introducing a new perspective to the wildlife re-identification field. In fact, preliminary attempts have been made to utilize Transformers to enhance high-frequency feature learning in person ReID tasks \cite{zhang2023pha}. Nevertheless, these efforts have not taken into account the influence of high-frequency noise in natural environments.

\begin{figure}[t]
    \centering
    \includegraphics[width=1\linewidth]{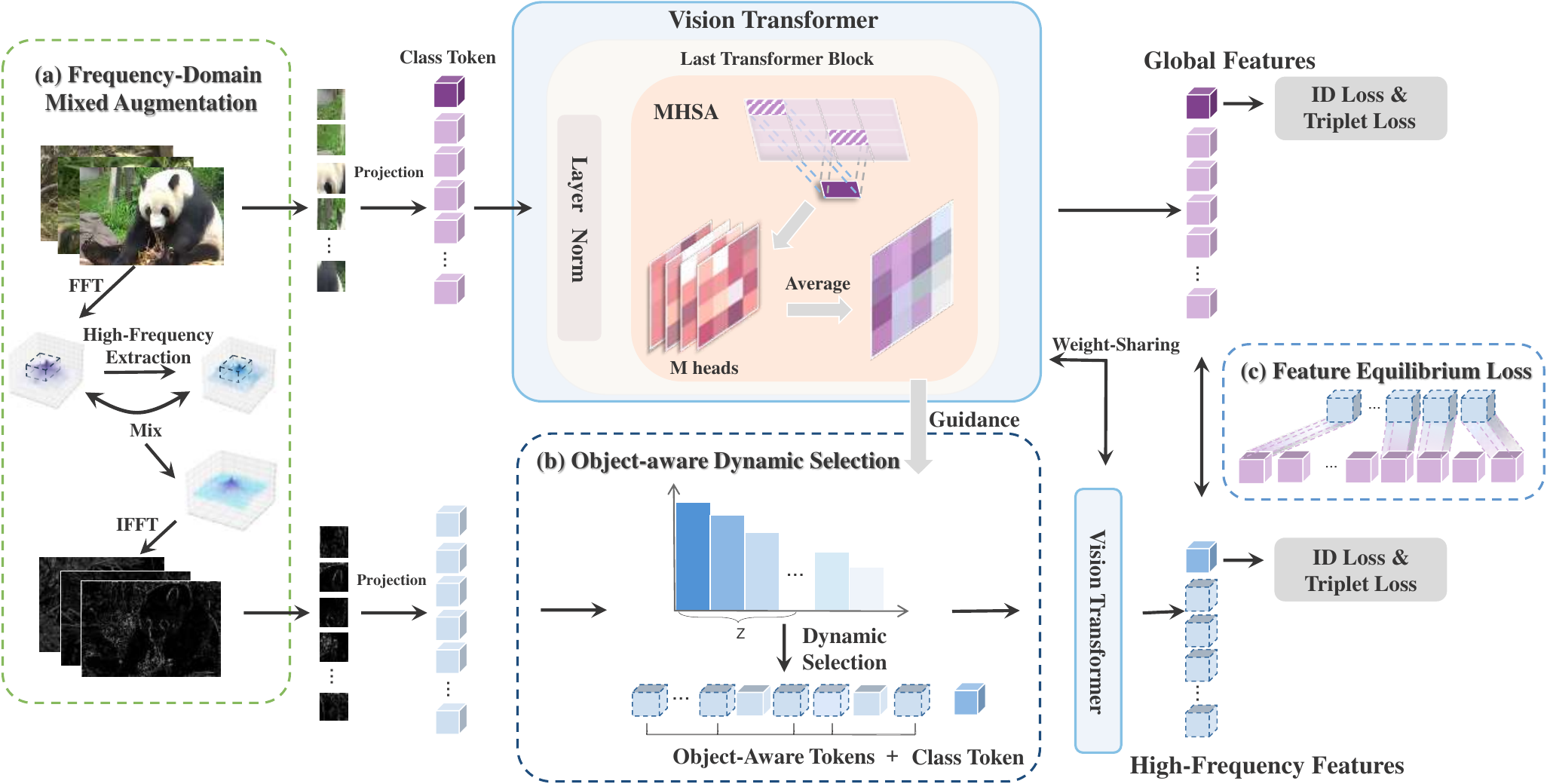}
    \caption{The architecture of our proposed method, consisting of (a) Frequency-Domain Mixed Augmentation(described in Sec.\ref{FMA}), (b) Object-Aware Dynamic Selection (described in Sec.\ref{OMS}), (c) Feature Equilibrium Loss (described in Sec.\ref{dgc}).  }
    \label{img1}
\end{figure}

\section{Method}
In this section, the specific details of the proposed method are presented. 
As shown in Fig.\ref {img1}, our adaptive high-frequency Transformer, with the core objective of enhancing high-frequency feature learning, incorporates three strategies. 1) Frequency-domain mixed augmentation. We introduce a novel data augmentation method by mixing high-frequency information and original information at the frequency domain level. This method specifically addresses the instability of high-frequency details resulting from variations in lighting and posture, contributing to improved model performance under diverse conditions.  2) Object-aware dynamic selection. We utilize global attention to flexibly mine high-frequency regions in images related to wildlife targets, facilitating the explicit learning of high-frequency features. This strategy allows us to minimize the influence of environmental noise on learning high-frequency features. 3) Feature equilibrium loss. In order to prevent excessive emphasis on high-frequency information from interfering with the original visual information during the feature learning process, we further propose a loss to balance their relationship.

\subsection{ViT ReID Baseline.}
\label{base}
Our model is built on the ReID baseline with vision transformer as backbone \cite{dosovitskiy2020image}. Given an image $I \in \mathbb{R}^{H \times W \times C}$, with height $H$, width $W$, and $C$ channels. The ViT model divides the image into $N$ fixed-size patches. These patches are then reshaped into a sequence of flattened vectors. The patches are linearly transformed into a $D$-dimensional embedding space. A learnable embedding, class token, is appended to this sequence for the purpose of capturing a global representation of the image, resulting in the sequence $\mathcal{X} \in \mathbb{R}^{(N+1) \times D}$. To capture the spatial information of the patches, positional embeddings $E_{pos}$ are introduced and combined with the patch embeddings, yielding the input $\mathcal{X} + E_{pos}$, where $E_{pos} \in \mathbb{R}^{(N+1) \times D}$. The final input to the transformer's encoder is thus a combination of patch embeddings, positional embeddings, and the class token. To optimize the model's parameters, a combination of loss functions is employed. After learning, the class token is further used as a global feature representation, denoted as $c$. The triplet loss, denoted by $\mathcal{L}_{tri}$, and the ID loss, denoted by $\mathcal{L}_{ID}$, are integral to the network optimization process for ReID tasks.

\subsection{Adaptive High-Frequency Transformer}
\subsubsection{Frequency-Domain Mixed Augmentation.}
\label{FMA}
For wildlife, high-frequency details such as fur patterns and small spots may be unstable due to variations in lighting and posture. In this part, we propose a data augmentation strategy named Frequency-Domain Mixed Augmentation(FMA), aimed at enhancing the robustness of models to simulate the detail changes caused by environmental factors such as seasonal fur variations or occlusions by mud, allowing the model to focus more on stable and essential features of the images. 
In brief, we transform the spatial representation of an image into a frequency domain and extract high-frequency information to obtain a representation dominated by high frequencies. The frequency domain representation of the original image is mixed with the high-frequency representation, thereby generating an augmented representation. Operating in the frequency domain is motivated by the recognition that each pixel manipulation in the spatial domain simultaneously adjusts multiple frequency components within the image. Mixing at the image level may introduce new high-frequency information, undermining the effective smoothing of high-frequency components. Specific steps are presented.

\label{ods}

\textbf{High-Frequency Information Extraction.} We first transform an input image $I\in\mathbb{R}^{H \times W \times C}$ into a single-channel image; each pixel in the image is located at coordinates \( (x, y) \) with a value \( I(x, y) \). The Fourier transformation is utilized to convert the spatial representation of the image into the frequency domain \( F(I) \):
\begin{equation}
F(I)(u, v) = \sum_{x=0}^{H-1} \sum_{y=0}^{W-1} I(x, y) e^{-j2\pi\left(\frac{ux}{H} + \frac{vy}{W}\right)}.
\end{equation}
This transformation \( F(u, v) \) maps the spatial information to the frequency domain, where \( u \) and \( v \) represent the frequency components in the horizontal and vertical directions, respectively. After processing the frequency domain representation with a Gaussian high-pass filter, we obtain a filtered version that contains only high-frequency components denoted by \( F_{h}(I) \). \par

\textbf{Frequency-Domain Mixing.} The FMA method blends the high-frequency components with the original image within the frequency domain, thereby sharpening the model's focus on stable features and improving its adaptability to environmental changes. We define a frequency mixing function, which randomly mixes \( F_{h}(I) \) with \( F(I) \):
\begin{equation}
F'_h{(I)}=(1-M_\alpha) \cdot F_h(I) + M_\alpha \cdot F(I).
\end{equation}
{$M_\alpha$} is a matrix of the same size as {$F_h{(I)}$} and {$F(I)$}, with a randomly selected square area covering $\alpha$(randomly ranging from 0 to 0.5) proportion of the total area set to 1, and the rest set to 0. The inverse Fourier transform of $F'_h{(I)}$ provides the augmented image for model training, to enhance feature stability recognition. Through this augmentation process, our model is more robust to the inevitable environmental factors of wildlife.
The augmented high-frequency representation, denoted by \( I_{h} \) and serving as the input high-frequency representation, typically represents finer details and edges within the image. \( I_{h} \) is derived by converting the augmented frequency domain representation $F'_h{(I)}$ back to the spatial domain. This inverse transformation is given by:
\begin{equation}
I_{h}(x,y) = \frac{1}{HW} \sum_{u=0}^{H-1} \sum_{v=0}^{W-1}F'_h{(I)}(u, v) e^{j2\pi\left(\frac{ux}{H} + \frac{vy}{W}\right)}.
\end{equation}

\subsubsection{Object-aware Dynamic Selection.}
\label{OMS}
High-frequency information reflects distinct discriminative features in images of various wild animals, demonstrating the versatility of high-frequency information across multiple species. However, the complex backgrounds in natural environments also fall under high-frequency information. Directly leveraging the high-frequency information of input images to enhance feature learning may result in excessive attention being focused on noise. 
Based on our frequency-domain mixed augmentation, we further introduce an object-aware high-frequency selection strategy that can adaptively adjust the extraction of high-frequency information to concentrate more on the target. Specifically, in the Vision Transformer's process of learning visual features from image inputs, the class token serves as a global aggregation, capturing holistic semantic information within the image. Particularly in ReID tasks, the class token plays a crucial role in directing the model's attention towards discriminative regions associated with the target. We leverage this global attention as guidance to selectively emphasize high-frequency patches of greater value, thereby enhancing the discriminative feature learning process. The strength of this strategy lies in its adaptability, as it is not restricted to specific species and effectively adapts to different wildlife targets. Detailed steps are as follows:

For an input image \(I\), we first divide it into the set  ${P}^o = \{{p}^o_i | i = 1,2,\ldots,n\}$, where $n$ represents the length of a patch sequence. Each patch ${p}^o_i$ is transformed into a high-dimensional embedding ${x}^o_i$ through a linear projection. Including the special class token embedding $x_{\text{[CLS]}}$, the set of all embeddings at the entry layer is denoted as ${\mathcal{X}^o = \{x^o_{\text{[CLS]}}, {x}^o_1,\ldots, {x}^o_n}$\}. At each layer $l$ of the ViT, the self-attention mechanism refines the embeddings based on inter-patch relationships. We designed a strategy based on attention to estimate the focus on the target:
\begin{equation}
\psi^l_{m,i} = \sigma \left( K^l_{m} x^l_i \odot Q^l_{m} x^l_{\text{[CLS]}} \right),
\end{equation}
where $\psi^l_{m,i}$ denotes the attention score of the \(i^{th}\) token relative to the [CLS] at the \(m^{th}\) head of the \(l^{th}\) layer. $\sigma$ represents the softmax function, which normalizes the computed attention scores. \(K^l_{m}\) is the transformation matrix for keys at the \(m^{th}\) head and \(l^{th}\) layer, and \(Q^l_{m}\) is the transformation matrix for queries. The $\odot$ represents the interaction between the \(i^{th}\) token and the [CLS]. 
Upon reaching the final layer \(L\), we compute the attention scores 
$ \Psi^L$, serving as a quantifiable metric that reflects the average attention distribution across heads in the model's final layer, defined as follows: 
\begin{equation}
\Psi^L = \frac{1}{M} \sum_{m=1}^{M} \psi^L_{m,i},
\end{equation}
where M represents the number of heads. This averaging process assists in revealing which parts of the input are given higher attention. $\Psi^L$ are analyzed to dynamically select the set of high-frequency information tokens that exhibit the highest attention scores, where the chosen tokens exhibit improved perceptual acuity towards the target. This selection is formalized as:
\begin{equation}
\mathcal{S}_{Z} = \{(\mathcal{O}(\Psi^L))_t | t = 1,2,\ldots,Z \},
\end{equation}
where $\mathcal{O}$ is a function that sorts scores in a set in descending order and then outputs the indices of these scores, $\mathcal{S}_{Z}$ represents the object perception token indices, $Z$ is computed by $\mu \cdot n$ and $\mu$ is a selection parameter. $\mathcal{S}_{Z}$ are stored in a dynamic memory $\mathcal{M}$, which later guides the dynamic selection process. 
 $\mathcal{X}^h =\{ {{x}^h_i | i = 1,2,\ldots,n}\}$ is the high-frequency information tokens counterpart of the original $\mathcal{X}^o$. To dynamically select tokens closely matching the target, we define a function $\Theta: \mathcal{S}_{Z} \rightarrow \mathcal{X}^h$ that selects the corresponding high-frequency tokens based on the indices determined by $\mathcal{S}_{Z}$. Thus, the input object-aware high-frequency embeddings are represented by ${\mathcal{X}^h{}' = \{x^h_{\text{[CLS]}}, {x}^h_1,\ldots, {x}^h_Z}$\}. Further, the global feature $c_o$ and $c_h$ are optimized with ID loss and triplet loss:
\begin{equation}
\mathcal{L} = \mathcal{L}_{ID}(c_o) + \mathcal{L}_{tri}(c_o) + \mathcal{L}_{ID}(c_h) + \mathcal{L}_{tri}(c_h).
\end{equation}

\subsection{Feature Equilibrium Loss}
\label{dgc}
Our model simultaneously takes visual image inputs and high-frequency augmented inputs, both of which are crucial for discriminative feature learning. The strategies we proposed above primarily guide the model to focus on high-frequency information. However, this needs to be established without compromising the learning of original visual information. Therefore, we further introduce the feature equilibrium loss to constrain the high-frequency features and visual features of the same individual from deviating excessively in the feature space.

Consider $f^h \in \mathbb{R}^{B\times Z \times D}$ denote the encoded high-frequency embeddings excluding class tokens, and $f^o \in \mathbb{R}^{B\times Z \times D}$ represent the encoded embeddings of the original sequence corresponding to $f^h$.
The proposed $\mathcal{L}_{F}$ aims to minimize the discrepancy between these two sets of embeddings, ensuring the retention of vital domain-specific features in the transformed embeddings.
Specifically, the feature equilibrium loss is defined as:
\begin{equation}
\mathcal{L}_{F} =  \sum_{b=1}^{B} \left( \frac{1}{Z} \sum_{z=1}^{Z} \|\vec{f}_{b,z}^o , \vec{f}_{b,z}^h \| \right).
\end{equation}
\( \|\vec{f}_{b,z}^o , \vec{f}_{b,z}^h \| \) represents the difference between the high-frequency feature \( \vec{f}_{b,z}^h \) and the original feature \( \vec{f}_{b,z}^o \) for the \( z \)-th token in the \( b \)-th input, detailed as:
\begin{equation}
\|\vec{f}_{b,z}^o , \vec{f}_{b,z}^h \| = \begin{cases} 
0.5 \left( \vec{f}_{b,z}^o - \vec{f}_{b,z}^h \right)^2, & \text{if } \left| \vec{f}_{b,z}^o - \vec{f}_{b,z}^h \right| < 1, \\
\left| \vec{f}_{b,z}^o - \vec{f}_{b,z}^h \right| - 0.5, & \text{otherwise}.
\end{cases}
\end{equation}
 Feature equilibrium loss aggregates the differences across all selected tokens, ensuring a comprehensive measure of the discrepancy between the high-frequency and original features for each token. By minimizing \(\mathcal {L}_{F} \), we encourage the model to preserve the essential features of the original input, while still leveraging the detailed textures and patterns enhanced in the high-frequency components, to ensure that the model learning does not overemphasize the high-frequency details at the expense of the original feature. This balance maintains visual and spatial consistency with the original feature while emphasizing high-frequency feature, thus improving the overall efficacy of feature extraction. The final loss function is denoted as:
\begin{equation}
\mathcal{L}_{\text{overall}} = \mathcal{L} + \lambda \mathcal{L}_{F},
\end{equation}
where $\lambda$ represents the weight of \(\mathcal {L}_{F} \).

\section{Experiments}
\subsection{Datasets and Evaluation Protocols}
\textbf{Datasets.} 
Existing studies for wildlife ReID exhibit inconsistent experimental settings regarding dataset partitioning, making it challenging to conduct fair comparisons between different approaches. Some methods only offer raw datasets without partition. In addition, most datasets are partitioned with varying standards, such as identity overlap between training and testing sets in some cases and no overlap in others.
To facilitate subsequent research in this field and provide a standardized data benchmark for related studies, we have divided the data into training and testing sets in a uniform manner, allocating 70\% of the identities for the training set and the remaining 30\% for the testing set, while ensuring no overlap between the identities in the training and testing sets. To test the universality of our method, We endeavored to encompass a diverse array of animal datasets, including species such as giant pandas\cite{wang2021giant}, elephants\cite{korschens2019elpephants}, seals\cite{nepovinnykh2022sealid}, giraffes\cite{parham2017animal}, sharks\cite{holmberg2009estimating}, tigers\cite{li2019atrw}, and pigeons\cite{kuncheva2022benchmark}. During the testing phase, each image in the test set is treated as a query, with the remainder of the test set, excluding the query image, forming the gallery. \par
\textbf{Evaluation Metrics.} In ReID tasks, two commonly used metrics are Cumulative Matching Characteristics (CMC) and mean Average Precision (mAP). We employed both CMC and mAP for evaluation. Notably, in wildlife ReID, most datasets lack explicit camera information, hence we include all correct matches in our evaluations. Given that in many instances, simple samples and minor viewpoint changes lead to higher Rank-k accuracy, we introduce a new metric, the mean Inverse Negative Penalty (mINP)\cite{ye2021deep}, which reflects the accuracy of identifying the most challenging matches.\par
\textbf{Implementation Details.} All of our experiments are conducted on PyTorch with Nvidia 3090 GPUs. We use the pre-trained vision transformer on imageNet-1K as the backbone. All images are resized to 256 $\times$ 256 and undergo data augmentation during training, which includes random rotations of 15 degrees, random adjustments to brightness and contrast with a 50\% probability each, and padding of 10 pixels. We configure $\mu$ to be 0.5 and $\lambda$ to be 0.1. The patch size is set to 16 $\times$ 16. During training, the SGD optimizer is used. The initial learning rate is 0.001, employing cosine learning rate decay. The training is conducted over 150 epochs with a batch size of 32, comprising 8 identities, each with 4 images. During the testing phase, distance matrices are computed using only the original features.

\begin{table}[ht]
\scriptsize
\renewcommand{\arraystretch}{1.3}
\centering
\caption{Comparison with the state-of-the-art methods on diverse wildlife datasets.}
\label{tab:swap}
\resizebox{\textwidth}{!}{%
\begin{tabular}{@{}l|ccc|ccc|ccc|ccc|ccc@{}}
\toprule
Method & \multicolumn{3}{c}{RotTrans\cite{chen2022rotation}} & \multicolumn{3}{c}{AGW \cite{ye2021deep}} & \multicolumn{3}{c}{TransReID\cite{he2021transreid}} & \multicolumn{3}{c}{CLIP-ReID\cite{li2023clip}} & \multicolumn{3}{c}{OURS} \\

\cmidrule(r){2-4} \cmidrule(lr){5-7} \cmidrule(lr){8-10} \cmidrule(lr){11-13} \cmidrule(l){14-16}
Dataset & mAP & R1 & mINP & mAP & R1 & mINP & mAP & R1 & mINP & mAP & R1 & mINP & mAP & R1 & mINP \\
\midrule
Panda\cite{wang2021giant} & 40.2 & 90.4 & 10.7 & 27.3 & 93.1 & 9.2 & 37.9 & 88.8 & 10.5 & 38.8 & 87.7 & 11.2 & \textbf{44.5} & \textbf{93.1} & \textbf{12.5} \\
Elephant\cite{korschens2019elpephants} & 29.1 & 54.1 & 9.3 & 21.9 & 48.7 & 5.1 & 21.2 & 50.9 & 4.1 & 20.4 & 43.7 & 5.3 & \textbf{30.4} & \textbf{58.0} & \textbf{9.3} \\
Seal\cite{nepovinnykh2022sealid} & 47.7 & 83.5 & 7.2 & 46.2 & 83.8 & 11.5 & 50.1 & 86.0 & 12.2 & 45.2 & 84.1 & 9.7 & \textbf{51.5} & \textbf{87.4} & \textbf{14.4} \\
Zebra \cite{parham2017animal} & 16.2 & 26.7 & 7.2 & 13.5 & 24.4 & 6.1 & 16.1 & 25.3 & 7.4 & 16.3 & 27.2 & 6.9 & \textbf{16.8} & \textbf{27.4} & \textbf{7.7} \\
Shark \cite{holmberg2009estimating} & 18.6 & 61.2 & 5.0 & 20.6 & 66.2 & 4.9 & 19.3 & 62.2 & 5.0 & 23.3 & 67.2 & \textbf{6.3} & \textbf{24.3} & \textbf{68.4} & 5.6 \\
Tiger \cite{li2019atrw} & 66.1 & 98.0 & \textbf{35.1} & 56.4 & 97.4 & 26.4 & 64.1 & 98.3 & 33.0 & 55.8 & 96.1 & 24.2 & \textbf{66.3} & \textbf{98.5} & 33.4 \\
Pigeon \cite{kuncheva2022benchmark} & 72.5 & 99.1 & 19.5 & 66.6 & 99.1 & 17.7 & 72.2 & 98.8 & 19.9 & 68.4 & 99.0 & \textbf{20.1} & \textbf{73.8} & \textbf{99.1} & 19.8 \\
Giraffe \cite{parham2017animal} & 48.4 & 46.7 & 39.9 & 44.2 & 43.5 & 35.6 & 45.8 & 42.4 & 37.5 & 47.6 & 43.5 & 39.5 & \textbf{49.1} & \textbf{47.8} & \textbf{40.2} \\
\bottomrule
\end{tabular}%
}
\end{table}

\begin{table}[htbp]
\centering
\begin{minipage}{0.48\linewidth}
\centering
\caption{Comparison in multi-species setting.}
\label{multi}
\resizebox{\textwidth}{!}{%
\begin{tabular}{@{}c|c|c c|c c|c c@{}}

\toprule
Method     & Reference  & \multicolumn{2}{c|}{Seal}  & \multicolumn{2}{c|}{Pigeon} & \multicolumn{2}{c}{Elephant} \\ 
           &            & mAP & R1 & mAP & R1 & mAP & R1 \\ 
\midrule
\makecell[c]{TransReID\cite{he2021transreid}}  & \makecell[c]{ICCV2021}    & 45.8& 82.6 &65.7  & 98.7 & 22.8  & 44.8 \\
\makecell[c]{CLIP-ReID\cite{li2023clip}}   & \makecell[c]{AAAI2023}     & 43.5 & 82.9 & 64.3 & 99.0 & 20.4 &44.2  \\
\makecell[c]{Ours}    & \makecell[c]{-} &  \textbf{50.6} &  \textbf{86.2} &  \textbf{70.0} &   \textbf{99.1 }& \textbf{26.6}   &   \textbf{52.6 }\\
\bottomrule
\end{tabular}%
}
\end{minipage}%
\hfill
\begin{minipage}{.48\linewidth}
\centering
\caption{Comparison in DG setting.}
\label{DG}
\resizebox{\textwidth}{!}{%
\begin{tabular}{@{}c|c|c c|c c|c c@{}}
\toprule
Method     & Reference  & \multicolumn{2}{c|}{AVG} & \multicolumn{2}{c|}{Tiger} & \multicolumn{2}{c}{Seal} \\ 
           &            & mAP & R1 & mAP & R1 & mAP & R1 \\ 
\midrule
\makecell[c]{CAL\cite{rao2021counterfactual}}    & \makecell[c]{ICCV2021}   & 42.8 & 58.0 & 64.1 & 63.4 & 21.6 & 52.7 \\
\makecell[c]{MetaBIN\cite{choi2021meta}}   & \makecell[c]{CVPR2021}   & 42.5 & 59.3 & 61.2 & 62.0 & 23.9 & 56.7 \\
\makecell[c]{UniReID\cite{jiao2024toward}}    & \makecell[c]{NIPS2023}   & 47.6 & 63.9 & 66.7 & 65.2 &  \textbf{28.5} & 62.6 \\
\makecell[c]{Ours}  & \makecell[c]{-} &  \textbf{48.1}  &  \textbf{88.5} &  \textbf{72.3} &  \textbf{98.1} & 24.0 &  \textbf{78.9} \\
\bottomrule
\end{tabular}%
}
\end{minipage}
\end{table}


The existing wildlife ReID experiments mostly employ different settings and do not provide publicly clear dataset divisions, making comparison impossible. Therefore, we compare our approach with the current state-of-the-art person ReID methods on our own divided dataset. Our model significantly outperforms existing ReID models based on CNN and ViT architectures, as shown in Table.\ref{tab:swap}. To demonstrate the universality of our model, we evaluate it on multiple wildlife datasets, including terrestrial, aquatic, and aerial species. Among these, species like elephants and sharks, which lack distinct pattern features, require emphasis on contour line recognition, whereas tigers and seals necessitate fine-grained pattern recognition.

\textbf{Comparison with the SOTA supervised ReID methods.} As shown in Table.\ref{tab:swap}, the overall Rank-1 accuracy is high. This outcome is partly due to the lack of camera information, with some images under the same camera experiencing minor perspective changes, making simple samples easily identifiable. The observed low mAP suggests a low level of recognition accuracy across the board, signifying the significant challenge presented by the task of wildlife ReID. Experiments show that ViT-based methods outperform the CNN-based methods on most datasets. Our model exhibits a superior performance on most datasets. Specifically, compared to the latest large-scale pre-trained CLIP-ReID, our model significantly outperforms it in wildlife ReID datasets.

\textbf{Comparison on Multi-Species ReID and DG settings.} The multi-species setting refers to training a combined dataset composed of training sets of elephants, seals, pigeons, and pandas, then testing on one of these species' test sets. Compared to single species, multi-species ReID requires a higher level of generalization and poses more significant challenges. It demands the careful balancing of learning among various species. Despite these challenges, our model demonstrates superior performance over existing methods within the multi-species setting, as presented in Table.\ref{multi}. 
For the domain generalization setting, the training set used is Wildlife-71\cite{jiao2024toward}, with testing conducted on tigers and pigeons datasets. We mainly compare with the SOTA DG method UniReID. Only the Tiger and Seal datasets in UniReID have publicly available partitioning methods, and the Tiger dataset in UniReID is different from the official version used in Table.\ref{tab:swap}, makes it incomparable to that result. To ensure a fair comparison, we adopt the same settings as UniReID in DG experiments. It is worth noting that, even without large-scale model pre-training and introducing test set images as guidance, our average mAP on both the tiger and seal datasets remains significantly higher than UniReID, as shown in Table.\ref{DG}, demonstrating the generalizability of our approach.

\subsection{Ablation Studies}

Ablation studies are performed on the panda and pigeon datasets to validate
the effectiveness of our method. We first compare it with the ViT baselines, described in Sec.\ref{base}. Then, we sequentially added our core designs to the baseline to test the improvements introduced by our design, as shown in Table.\ref{ablation}.



\begin{table}[t]
\scriptsize
\centering
\renewcommand{\arraystretch}{1.3}
\caption{Ablation study on several datasets. mAP is reported.}
\label{ablation}
\begin{tabular}{lccccc}
\toprule
Strategy & Panda & Pigeon & Giraffe & Shark & Seal\\
\midrule
Baseline (ViT)\cite{dosovitskiy2020image} & 40.8 & 70.1 & 47.5 & 20.2 & 49.5 \\ 
Pure High-Frequency Augmentation & 41.8 & 68.4 & 47.4 & 21.5 & 50.1 \\
PHA(reproduced) \cite{zhang2023pha} &38.8 & 70.7 & 47.5 & 14.8 & 47.7 \\
\hline
\multicolumn{5}{l}{\textit{Adaptive High-Frequency Transformer}} \\ 
\hline
+ Frequency-Domain Mixed Augmentation & 42.7 & 70.9 & 47.8 & 21.7 & 50.8 \\
+ Object-aware Dynamic Selection & 43.9 & 73.6 & 48.7 & 23.6 & 51.3 \\
+ Feature Equilibrium Loss & 44.5 & 73.8 & 49.1 & 24.3 & 51.5 \\
\bottomrule
\end{tabular}
\end{table}

\begin{figure}
    \centering
    \includegraphics[width=1\linewidth]{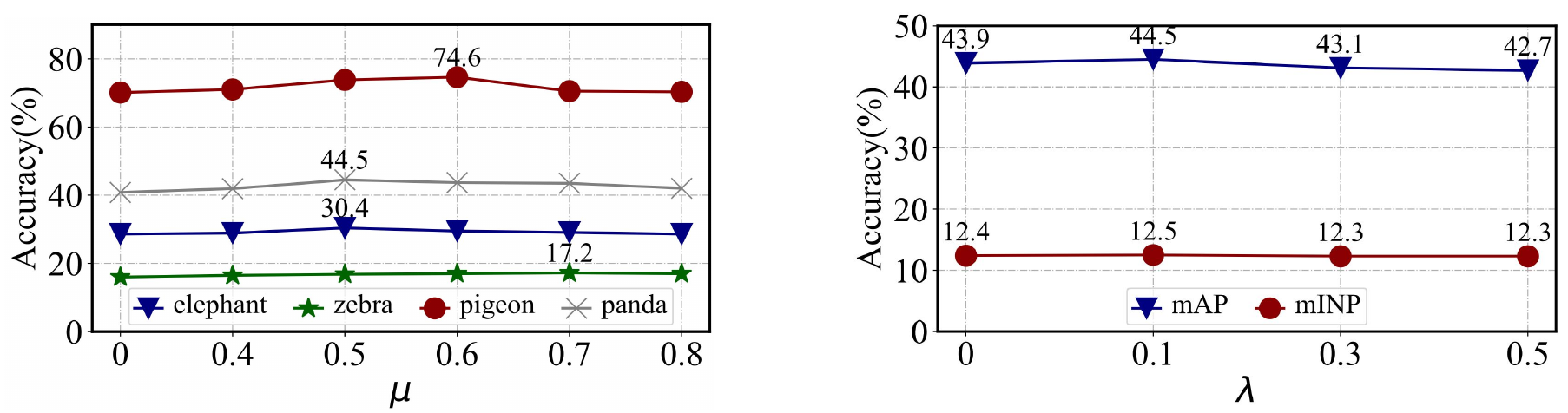}
    \caption{Parameter evaluation. mAP results for varying $\mu$ are compared across several datasets. Different weights $\lambda$ are analyzed on the Panda dataset.}
    \label{parameter}
\end{figure}

\begin{figure}[t]
    \centering
    \includegraphics[width=1\linewidth]{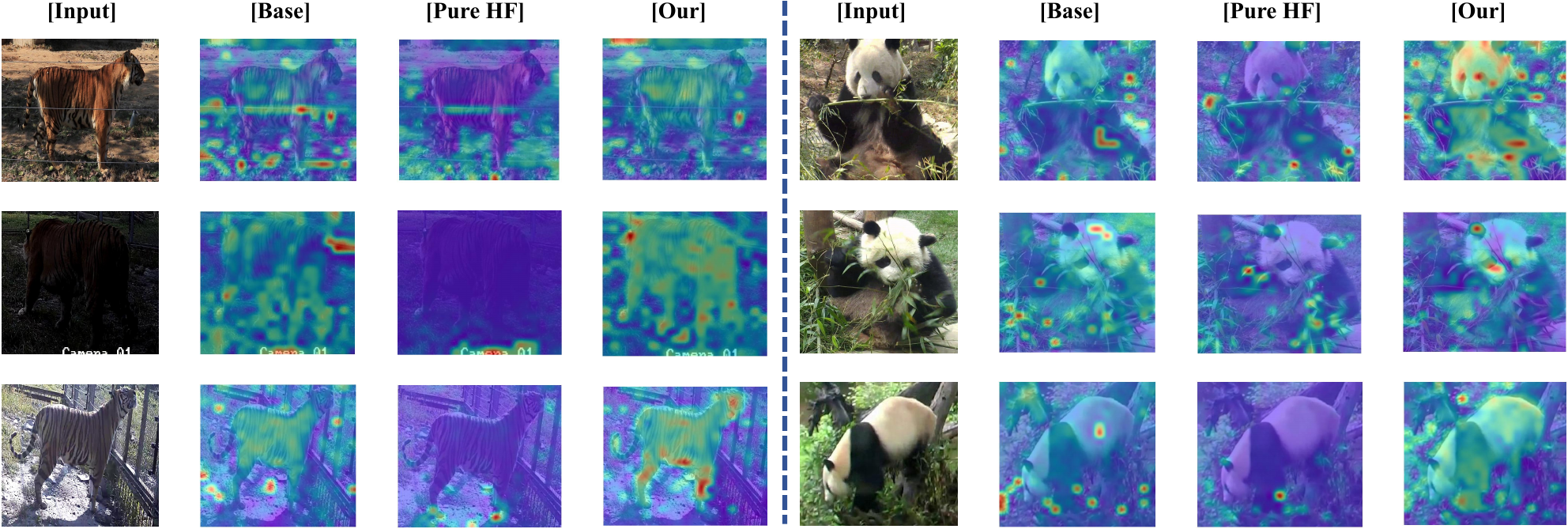}
    \caption{Visualizing the attention maps for the class token from the last self-attention layer. [Base] denotes the baseline. [Pure HF] refers to the Pure High-Frequency Augmentation as detailed in Table.\ref{ablation}. [Our] means the method we proposed in this paper.}
    \label{attn}
\end{figure}

\textbf{Pure High-Frequency Augmentation.} To validate the effectiveness of enhancing high-frequency feature learning for wildlife ReID, a straightforward approach involves directly extracting high-frequency information from the images for augmentation. We conducted relevant experiments in this part. In Table.\ref{ablation}, Pure High-Frequency Augmentation refers to training exclusively with the high-frequency information of the original images without undergoing our frequency-domain mixed augmentation operations. Compared to the baseline, our method shows improvements in mAP, Rank1, and mINP on the panda dataset. While on the pigeon dataset, the results showed that the mAP declined by 1.7\% compared to the baseline. The results demonstrate that although high-frequency information plays a crucial role in enhancing wildlife contours and textures, it also intensifies high-frequency background noise. This leads to a less noticeable performance improvement and, in some cases, a decline. Subsequent experiments and visualization analyses also confirm this, shown in Fig.\ref{attn}.

\textbf{The Effectiveness of Our Methods.} \textit{1) Frequency-Domain Mixed Augmentation.} We conducted experiments to verify that our frequency domain mixing augmentation enhances model robustness. This approach emphasizes the model's reliance on stable features, such as general body shape, while maintaining sensitivity to high-frequency details, despite environmental factors that may obscure high-frequency information. Experimental results show a 0.9\% increase in mAP and a 0.2\% increase in mINP on the panda dataset compared to Pure High-Frequency Augmentation, with improvements also observed in the pigeon dataset. \textit{2) Object-aware Dynamic Selection.} To confirm that our Object-aware Dynamic Selection (ODS) method can more effectively learn high-frequency target information and reduce background interference, we continued testing with Frequency-Domain Mixed Augmentation. Experiments validating the effectiveness of ODS demonstrate increased mAP on multiple animal datasets compared to the Baseline. Additionally, visualized attention maps indicate that ODS enables the model to focus more on the target than both the Baseline and Pure High-Frequency Augmentation. Compared to PHA \cite{zhang2023pha}, which we have reproduced to closely resemble the source version, our ODS demonstrates superior performance. PHA amplifies uncertain local high-frequency features, which may lead to a bias toward background noise. In contrast, our method extracts object-aware high-frequency information and reduces background noise. \textit{3) Feature Equilibrium Loss.} We conducted experiments to compare the model's performance with and without feature equilibrium loss. Feature equilibrium loss balances original and high-frequency features, reducing their disparity. The results show that this balance enhances nuanced consistency.

\textbf{Parameter Analysis.} We conduct a thorough evaluation of the effects exerted by the ratio $\mu$ across several datasets and the feature equilibrium loss weight $\lambda$ on the panda dataset. The experimental outcomes depicted in Fig.\ref{parameter} elucidate the feature equilibrium loss performs better when assigning lower weights, with weight 0.1 the best. The optimal value of $\mu$ varies across different datasets. Due to the absence of bounding boxes, the proportion of the target within the entire image varies across different datasets. This variation can lead to the selection of too few targets or excessive background when adjusting $\mu$.


\textbf{Visualization Analysis.} Fig.\ref{attn} exhibits the attention maps of the Baseline, Pure High-Frequency Augmentation(Pure HF), and our model. It is evident that the Baseline attention significantly surpasses that of Pure HF in terms of focus on the target object. Pure HF is considerably affected by background noise, resulting in less attention to the target. In contrast, our model is capable of not only locating the target but also focusing on the high-frequency regions of the target, effectively enhancing the recognition of contours and textures.

\section{Conclusion}
In this paper, we analyze the challenges unique to wildlife ReID compared to conventional ReID tasks and propose a versatile adaptive high-frequency Transformer architecture tailored to achieve effective performance across diverse wildlife species. Specifically, we propose enhancing feature learning by focusing on high-frequency information that can capture the distinct characteristics of various animals. Extensive experimental evaluations on different wildlife species demonstrate the state-of-the-art performance of our model in the ReID task. Experiments under domain generalization settings also showcase the generalization capability of our model to unknown species. \par

\textbf{Limitations.} This part discusses the limitations of our method. It is somewhat influenced by the baseline choice due to its reliance on the dynamic selection process based on baseline attention, meaning poor baseline attention could lead to selecting high-frequency information with more background noise. Besides, the potential for variability in the optimal selection of the value of $\mu$ across different datasets suggests that our approach may not achieve complete adaptability. In the future, we will attempt to design more flexible strategies to dynamically adjust the value of $\mu$ according to the specific features of each dataset.

\bigskip

\noindent\textbf{Acknowledgments.}
This work is supported by the National Natural Science Foundation of China under Grant (62176188, 62361166629) and the Special Fund of Hubei Luojia Laboratory (220100015). The numerical calculations in this paper have been done on the supercomputing system in the Supercomputing Center of Wuhan University.

\clearpage  

%
%
\bibliographystyle{splncs04}
\bibliography{main}
\end{document}